\documentclass[preprint,12pt]{elsarticle}

\usepackage{amssymb}
\usepackage{amsmath}
\usepackage{graphicx}
\usepackage[colorlinks=true, allcolors=blue]{hyperref}
\usepackage[utf8]{inputenc}
\usepackage{setspace}
\usepackage{subcaption}
\usepackage{multirow}
\usepackage{lineno}
\usepackage{hyperref}
\usepackage{booktabs}
\usepackage{float}
\usepackage{makecell}
\usepackage{amsfonts}
\usepackage{bm}
\usepackage{longtable}
\usepackage{subcaption}
\usepackage{orcidlink}

\journal{Informatics in Medicine Unlocked}

\begin{document}

\begin{frontmatter}



\title{PAL-Net: A Point-Wise CNN with Patch-Attention for 3D Anatomical Facial Landmark Localization} 

\author[inst1]{Ali Shadman Yazdi \corref{cor1}\orcidlink{0009-0002-8317-8332}} 
\author[inst2,inst3]{Annalisa Cappella \orcidlink{0000-0002-4527-4203}} 
\author[inst1]{Benedetta Baldini \orcidlink{0000-0003-4365-4998}} 
\author[inst2]{Riccardo Solazzo \orcidlink{0009-0006-0085-3187}} 
\author[inst4,inst5]{Gianluca Tartaglia \orcidlink{0000-0001-7062-5143}} 
\author[inst2]{Chiarella Sforza \orcidlink{0000-0001-6532-6464}} 
\author[inst1]{Giuseppe Baselli  \orcidlink{0000-0003-2978-1704}} 

\cortext[cor1]{Corresponding author. 
  \textit{email address:} ali.shadman@polimi.it
  \textit{address:} Via Camillo Golgi, 39, 20133 Milano MI
  \textit{Phone number:} +393669094310}

\address[inst1]{Department of Electronics, Information and Bioengineering, Politecnico di Milano, Milan, Italy.}

\address[inst2]{Department of Biomedical Sciences for Health, University of Milan, Milan, Italy.}

\address[inst3]{U.O. Laboratory of Applied Morphology, IRCCS Policlinico San Donato, 20097 San Donato Milanese, Italy.}

\address[inst4]{Department of Biomedical, Surgical and Dental Sciences, University of Milan, Milan, Italy.}

\address[inst5]{Fondazione IRCCS Cà Granda, Ospedale Maggiore Policlinico, 20122 Milan, Italy.}

\begin{abstract}

Manual annotation of anatomical landmarks on 3D facial scans is a time-consuming and expertise-dependent task, yet it remains critical for clinical assessments, morphometric analysis, and craniofacial research. While several deep learning methods have been proposed for facial landmark localization, most focus on pseudo-landmarks or require complex input representations, limiting their clinical applicability. This study presents a fully automated deep learning pipeline (PAL-Net) for localizing 50 anatomical landmarks on facial models acquired via stereo-photogrammetry. The method combines coarse alignment, region-of-interest filtering, and an initial landmark approximation with a patch-based pointwise CNN enhanced by attention mechanisms. Trained and evaluated on 214 annotated scans from healthy adults, PAL-Net achieved a mean localization error of 3.686 mm and preserved relevant anatomical distances with an average error of 2.822 mm. While the geometric error exceeds expert intra-observer variability, the distance-wise error maintains structural integrity sufficient for high-throughput anthropometric analysis. To assess generalization, the model was further evaluated on 700 subjects from the FaceScape dataset, achieving a mean localization error of  0.41\,mm and a distance error of 0.38\,mm. Comparing with existing methods, PAL-Net offers a favorable trade-off between accuracy and computational cost. While performance degrades in regions with poor mesh quality (e.g., ears, hairline), the method demonstrates consistent accuracy across most anatomical regions. PAL-Net generalizes effectively across datasets and facial regions, outperforming existing methods in both point-wise and structural evaluations. It provides a lightweight, scalable solution for high-throughput 3D anthropometric analysis, with potential to support clinical workflows and reduce reliance on manual annotation. Source code can be accessed at https://github.com/Ali5hadman/PAL-Net-A-Point-Wise-CNN-with-Patch-Attention. 

\end{abstract}

\begin{graphicalabstract}
\begin{figure}[!h]
\centering
\includegraphics[width=0.79\linewidth]{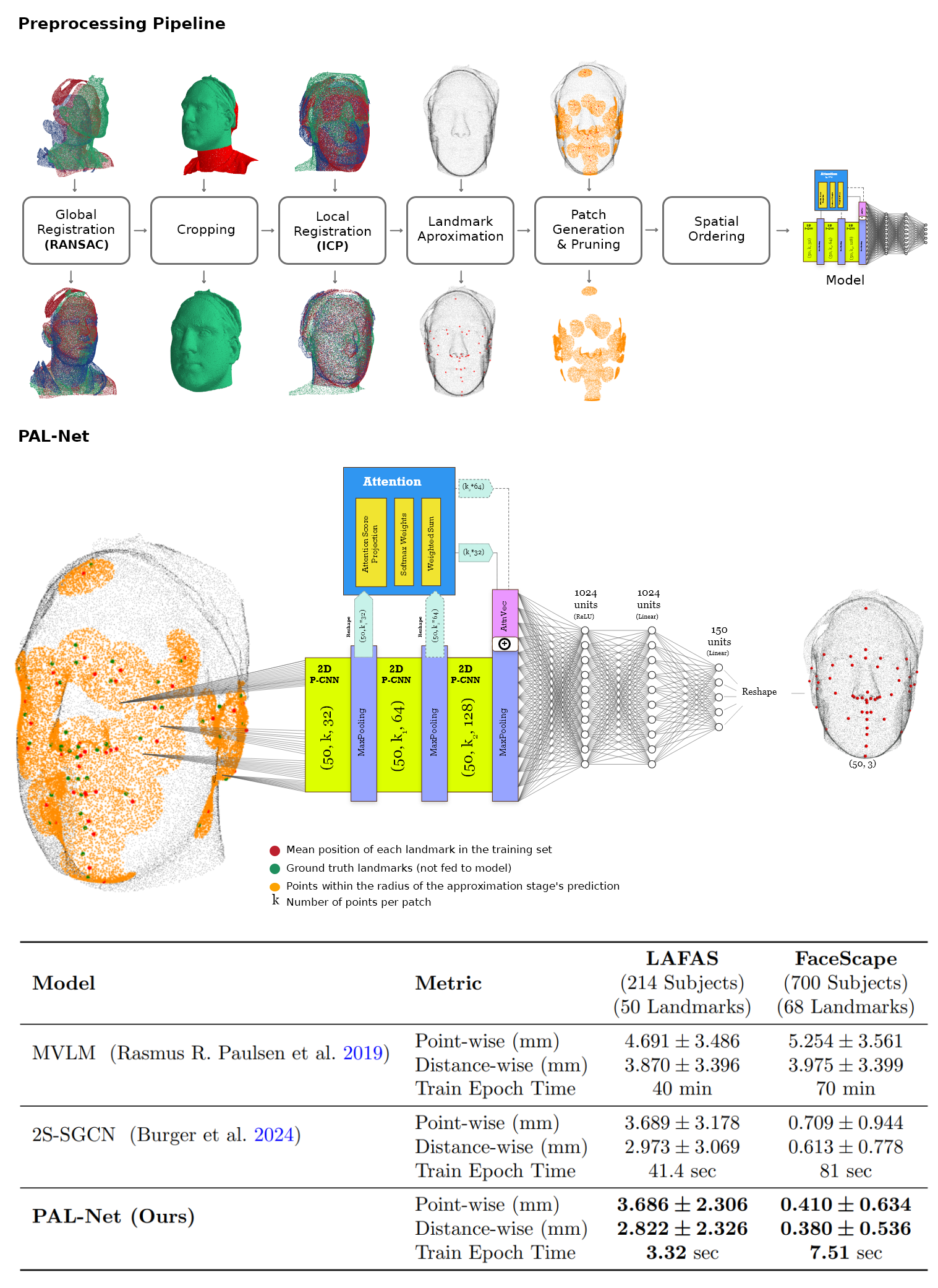}
\end{figure}

\end{graphicalabstract}

\begin{highlights}
\item PAL-Net model for 3D anatomical landmarking using patch point-wise CNNs
\item Predicts 50 anatomical landmarks with 3.68 mm error on stereo-photogrammetric facial scans
\item Global attention improves global feature learning and structural accuracy
\item PAL-Net outperforms prior methods with lower training time and memory usage
\item Generalizes well to external data, enabling fast, accurate 3D facial analysis

\end{highlights}

\begin{keyword}
3D facial landmark localization \sep Anatomical landmarks \sep Patch-based CNN \sep Point-wise convolution \sep Facial morphometrics \sep 


\end{keyword}

\end{frontmatter}



\section{Introduction}

Precise localization of anatomical facial landmarks is fundamental to clinical craniofacial analysis, guiding surgical planning, outcome assessment, and dysmorphology diagnosis \citep{masnada20203d, chang2015three, codari2017computer, see_age-_2008}. Beyond pathology, 3D anthropometry is essential for studying growth trajectories and population variations \citep{gibelli2022three}. While modern stereo-photogrammetry systems deliver high-resolution 3D meshes comparable to those from laser scanning \citep{heike_3d_2010}, downstream analysis still rely heavily on accurate landmarking. However, facial landmarks fall into two distinct categories: anatomical and pseudo-landmarks. While pseudo-landmarks are algorithmically defined by geometric or statistical properties of the facial images and widely used in general computer vision, anatomical landmarks are standard for medical anthropometry. They correspond to specific biological structures \citep{katina2016definitions} and have consistent definitions across different individuals \citep{facchi2025graph}.

Despite the advancements in digitized anthropometry, identifying these anatomical points remains a bottleneck. Manual annotation is labor-intensive and prone to inter-observer variability, particularly in regions with few distinct geometric features \citep{gibelli2020reliability, fagertun20143d}. Existing automated methods often focus on pseudo-landmarks or require computationally expensive inputs, limiting their direct clinical applicability. To address this, we introduce PAL-Net, a deep learning framework tailored for the efficient and accurate localization of 50 anatomical landmarks on 3D facial scans. The main contributions of this study are summarized as follows:

\begin{itemize}
    \item We propose PAL-Net, a lightweight patch-based point-wise CNN that integrates a global attention mechanism to capture both local geometry and global facial structure.

    \item We introduce a composite loss function that explicitly penalizes deviations in inter-landmark distances. This structure-preserving design ensures that the predicted geometry maintains the anthropometric consistency required for clinical measurements.

    \item We design the framework to operate directly on raw, unstructured point clouds, eliminating the need for computationally intensive preprocessing.
    
    \item We target a comprehensive set of 50 clinically validated anatomical landmarks overcoming the limitations of previous methods restricted to pseudo-landmarks.
    
\item We demonstrate that PAL-Net achieves a favorable trade-off, providing accuracy that is sufficient for anthropometric use while significantly reducing computational cost compared to state-of-the-art methods.
    
    \item We provide a robust evaluation on both a clinically annotated dataset (LAFAS) and a large-scale public dataset (FaceScape), validating the model's generalization capabilities.
\end{itemize}

The remainder of this paper is organized as follows. Section 2 outlines the problem formulation and details the proposed methodology, including data preprocessing and the PAL-Net architecture. Section 3 describes the datasets, training protocols, and experimental setup. Section 4 presents the comparative and anatomical performance analysis, including ablation studies. Finally, Section 5 discusses limitations and concludes the study with directions for future work.

\subsection{Related Works}

Approaches for 3D anatomical facial landmark localization have progressed from classical geometric algorithms to modern learning-based techniques. Initially, traditional methods, predating machine learning, focused on hand-crafted geometric feature extraction and model fitting. These techniques identify landmarks using local surface properties such as principal curvatures, shape indices, or geodesic distances \citep{tang2017principal}. Additionally, some methods employ statistical shape models, like 3D Active Shape Models (ASMs), or use deformable models to fit a template to a target \citep{frantz2000localization}. In parallel, dense correspondence methods attempt to map the entire facial surface by modeling coherent local deformations; for example, boosting local shape matching \citep{fan2019boosting} and landmark-free registration \citep{fan2023landmark}. While these dense correspondence techniques can implicitly localize landmarks via full-template alignment, they address a broader and more computationally intensive registration problem. In contrast, our method focuses on directly and efficiently localizing a sparse set of clinically relevant anatomical points, optimizing for high-throughput anthropometric analysis.

To address the need for efficiency and robustness, recent studies have investigated the application of machine learning (ML) and deep learning (DL) techniques specifically for landmark localization. Since pseudo-landmarks are algorithmically determined, they exhibit high consistency with minimal variability; therefore, most studies on facial landmark localization develop their models using these landmarks, as their consistency often leads to improved results. \citet{creusot2013machine} used machine learning to annotate 14 facial landmarks on 3D models by manually extracting features such as vector normals, neighboring point data, and principal curvatures. Their approach, combining offline training and online detection, achieved prediction errors ranging from 2.5 to 10 mm on non-anatomical landmarks. \citet{o2019extending} demonstrated the feasibility of adapting 2D methods to 3D facial landmark localization by extending Convolutional Pose Machines (CPMs) \citep{wei2016convolutional} for 3D facial landmark localization, by refining heat-maps using PointNet++ \citep{qi2017pointnet++} architecture. Their approach demonstrates the feasibility of extending 2D methods to 3D facial landmark localization. \citet{paulsen_multi-view_2019} achieved a localization error of 2.42mm using a multi-view approach for facial landmark identification on the BU-3DFE datasets \citep{yin20063d}, comprising 83 non-anatomical facial landmarks. Although the results are very accurate, the methodology requires significantly high computational power. It involves rendering the 3D facial model from multiple views, generating multiple 2D images of the face, and applying 2D methods for facial landmark localization. \citet{wang2022learning} used Graph Convolutional Networks (GCN) with PAConv \citep{xu2021paconv} for 3D facial landmark localization, learning self-attention, and refining affine transformations. Their method achieved a 2mm localization error, outperforming others on the same dataset. \citet{burger20242s} employed a two-stage stratified graph convolutional network model for facial landmark detection on 3D data(2S-SGCN). The first stage uses a stratified graph convolutional network (SGCN) to detect landmark regions by combining global and local graph representations. The second stage refines these predictions with the MSE-over-mesh method to accurately locate landmarks directly on the mesh. This approach, while being computationally efficient, achieves an average localization accuracy of approximately 0.371mm on 3D scans from the FaceScape dataset \citep{yang2020facescape}.

The mentioned studies have applied their methodology to pseudo-landmarks due to their consistency. However, fewer studies have applied various methods to localize anatomical landmarks, due to the limited availability of annotated training data, especially in medical imaging contexts \citep{zhang2017detecting}. This scarcity makes it difficult to train models effectively, leading to lower accuracy in landmark detection. \citet{chong2024automated} developed a U-NET-based deep learning algorithm \citep{ronneberger2015u} for automated anatomical landmark detection on 3D facial images. The method involves stacking two U-NETs for coarse and fine feature extraction, followed by back-projection to accurately localize 20 landmarks. The model was validated on healthy subjects, acromegaly patients, and localized scleroderma patients, achieving an average normalized mean error (NME) of 1.4 mm for healthy cases, 2.2 mm for scleroderma patients, and 2.8 mm for acromegaly patients. \citet{guo2013automatic} developed a method for automatic 3D facial landmark annotation using a combination of anatomical landmarks and pseudo-landmarks for automatic facial image registration. They began with the manual annotation of six key anatomical landmarks, which were then automatically localized using PCA-based feature recognition. In addition to these anatomical landmarks, 11 pseudo-landmarks were heuristically identified based on geometric relations and texture constraints. The study reported that the localization error for automatic landmark annotation ranged between 1.0 mm and 3.6 mm with factors like facial hair affecting the accuracy in some regions. \citet{berends2024fully} introduced a fully automated pipeline for localizing ten soft-tissue landmarks on 3D facial meshes using a two-stage DiffusionNet architecture. Their method combines spectral surface features (HKS), pose normalization, and non-rigid registration to achieve high accuracy (1.69 ± 1.15 mm) on a dataset of 2,897 subjects. More recently, sophisticated deep learning architectures such as Transformer-based architectures have been explored for 3D position and anatomical landmark localization. For instance, \citet{zhang20263d} proposed a cascade Point Transformer for landmark localization on 3D human scans, while \citet{kubik2024leveraging} applied a Point Transformer module for anatomical landmark detection on 3D dental images. These approaches, while powerful,often introduce significant computational complexity, motivating the need for more lightweight solutions. Addressing perior studies' limitations such as manual feature extraction, limited anatomical scope, and high computational requirements, this work proposes a lightweight deep learning-based approach to fully automate 3D anatomical facial landmark annotation. While existing methods often focus on a small subset of identifiable anatomical landmarks \citep{manal2019survey}, typically focusing on fewer than 20, our approach targets 50 clinically validated anatomical landmarks across diverse facial regions, including areas known to be more challenging to localize. The proposed model was trained and evaluated on a dataset of 214  stereo-photogrammetric facial models with expert-annotated anatomical landmarks. The proposed system uses minimal preprocessing and operates directly on point clouds without expensive mesh fitting or spectral descriptors. The methodology is scalable for real-world clinical and research settings, aiming to reduce manual annotation time, improve reproducibility, and provide a standardized high-resolution tool for both clinical and research applications in 3D anthropometric analysis.

\section{Methods}

\begin{figure}
\centering
\includegraphics[width=\linewidth]{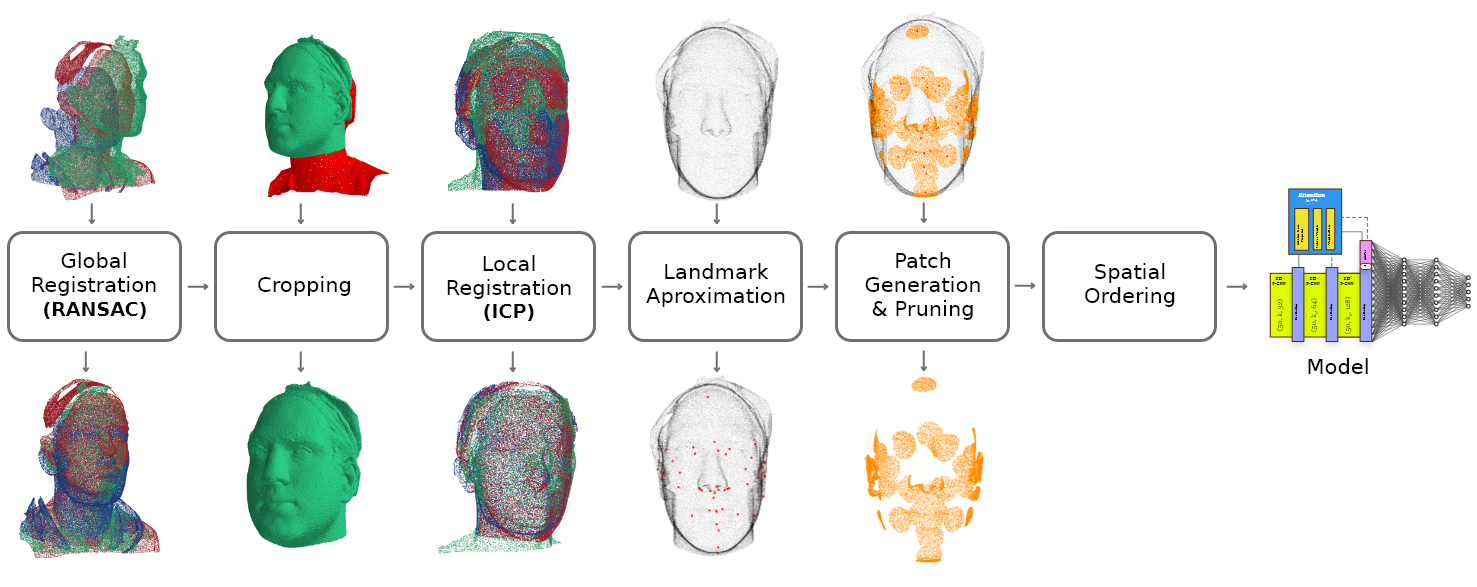}
\caption{\label{fig:preprocess} Preprocessing pipeline for 3D facial data used in the study. The pipeline consists of coarse registration, cropping, local registration.}
\end{figure}

This section describes the proposed pipeline for fully automated anatomical landmark localization on 3D facial meshes acquired via stereo-photogrammetry. The approach is developed and evaluated using the LAFAS dataset, a curated collection of 3D facial scans annotated with 50 clinically validated anatomical landmarks (see Section \ref{datasets} for details). The main stages of the pipeline include: preprocessing and rigid alignment, coarse landmark approximation via population-based templates, prediction using a patch-based point-wise CNN with attention (PAL-Net), and evaluation based on both point-wise and inter-landmark distance errors. An overview of the preprocessing steps is shown in Figure~\ref{fig:preprocess}. 

\subsection{Preliminaries}

Let a 3D facial model be represented as a point cloud $\mathcal{F} = \{\mathbf{f}_i\}_{i=1}^{n_f} \subset \mathbb{R}^3$, where $n_f$ is the total number of vertices in the scan and each vertex $\mathbf{f}_i$ corresponds to its Cartesian coordinates $(x, y, z)$. The objective is to predict the 3D coordinates of $n_l$ predefined anatomical landmarks, for which the ground truth locations are denoted as the set $\mathcal{L} = \{\mathbf{l}_k\}_{k=1}^{n_l} \subset \mathbb{R}^3$, where each $\mathbf{l}_k$ is a point on the facial surface, $\mathcal{L} \subset \mathcal{F}$. Our model, PAL-Net, learns a mapping function $M$ to produce a set of predicted landmarks, $\hat{\mathcal{L}} = M(\mathcal{F})$, that minimizes a loss function $\mathcal{J}(\mathcal{L}, \hat{\mathcal{L}})$. To achieve this, our pipeline first generates an initial coarse approximation of the landmarks, $\bar{\mathcal{L}}_{\text{fit}} \in \mathbb{R}^{n_l \times 3}$, via atlas-based alignment. From these approximations, we extract $n_l$ local patches, where each patch $\mathcal{P}_k \in \mathbb{R}^{K \times 3}$ consists of the $K$ nearest points to its corresponding approximated landmark. The final input to our network is the collection of all patches, $\mathcal{P} = \{\mathcal{P}_1, \dots, \mathcal{P}_{n_l}\}$. The model then learns to refine these patches to predict the final, accurate landmark locations $\hat{\mathcal{L}}$.

\subsection{Data Preprocessing}

First, to ensure computational efficiency without sacrificing anatomical accuracy, each 3D facial model was resampled using a rejection-sampling strategy. This resulted in a low-resolution representation containing approximately 10,000 evenly distributed surface points, denoted as $\mathcal{F}_{\text{src}}^{\text{low}}$. Consequently, our pipeline utilizes two representations for each subject: the derived $\mathcal{F}_{\text{src}}^{\text{low}}$ for robust global registration, and the original high-resolution mesh ($\mathcal{F}_{\text{src}}^{\text{og}}$, $\approx$ 100,000 vertices) for detailed local alignment. To ensure consistent anatomical alignment, we implemented a multi-stage registration strategy using these two resolutions. We began with a global registration step using Fast Point Feature Histograms (FPFH) and RANSAC to align the source model $\mathcal{F}_{\text{src}}^{\text{low}}$ against a fixed reference template, denoted as $\mathcal{M}^{\text{low}}$. This step computes a coarse rigid alignment, producing a transformation matrix $\mathbf{T}_{\text{coarse}} $ , representing rigid body transformations within the Special Euclidean group $SE(3)$.

\[
\mathbf{T}_{\text{coarse}} \in SE(3), \quad \text{such that} \quad \mathcal{F}_{\text{aligned}}^{\text{low}} = \mathbf{T}_{\text{coarse}} \cdot \mathcal{F}_{\text{src}}^{\text{low}}
\]

This transformation is then applied to the original full-resolution facial mesh to bring it into the same coarse-aligned coordinate frame:
\[
\mathcal{F}_{\text{aligned}}^{\text{og}} = \mathbf{T}_{\text{coarse}} \cdot  \mathcal{F}_{\text{src}}^{\text{og}}
\]

The result provided a coarse alignment, which are then applied to the high-resolution models, yielding an anatomically consistent set of high-resolution facial models. To enhance subsequent local registration, irrelevant background geometry was removed using a Region-of-Interest (ROI) filter. Defined as a bounding box $\mathcal{B}_{\text{ref}} \subset \mathbb{R}^3$ on the reference, the filter cropped the mesh by discarding vertices $\mathbf{f}_i \notin \mathcal{B}_{\text{ref}}$, yielding the subset:
\[
\mathcal{F}_{\text{ROI}} = \left\{ \mathbf{f}_i \in \mathcal{F}_{\text{aligned}}^{\text{og}} \;\middle|\; \mathbf{f}_i \in \mathcal{B}_{\text{ref}} \right\}
\]

Cropping was performed at this stage solely to optimize the registration process, as the presence of irrelevant regions would have negatively affected both its accuracy and efficiency.  For fine alignment, a rigid transformation $\mathbf{T}_{\text{ICP}} \in SE(3)$ was estimated via Iterative Closest Point (ICP) between $\mathcal{F}_{\text{ROI}}$ and the reference, ensuring submillimeter correspondence. The final transformation composes the coarse and fine alignments, ensuring that no anatomical information was lost due to the temporary cropping.
\[
\mathbf{T}_{\text{final}} = \mathbf{T}_{\text{ICP}} \cdot \mathbf{T}_{\text{coarse}}
\]

Following registration, the dataset was partitioned into a training set (80\%) and a validation set (20\%) to balance data availability and evaluation needs. Given the limited sample size, no separate test set was created. 

To establish a population-based atlas, we computed the mean landmarks $\bar{\mathcal{L}}_{\text{mean}}$ by averaging the training set of $N$ samples: $\bar{\mathcal{L}}_{\text{mean}} = \frac{1}{N} \sum_{i=1}^{N} \mathcal{L}^{(i)}$. This template was then transferred to each individual mesh $\mathcal{F}_{\text{aligned}}^{\text{og}}$ to provide an initial landmark estimate. Specifically, each template landmark $\bar{\mathbf{l}}_k \in \bar{\mathcal{L}}_{\text{mean}}$ was projected onto the mesh surface by selecting the nearest vertex in Euclidean space i.e., $\bar{\mathbf{l}}_k^{\text{fit}} = \arg\min_{\mathbf{f}_i \in \mathcal{F}_{\text{aligned}}^{\text{og}}} \|\bar{\mathbf{l}}_k - \mathbf{f}_i\|_2$. The resulting set of fitted estimates, $\bar{\mathcal{L}}_{\text{fit}} \in \mathbb{R}^{n_l \times 3}$, serves as an initial approximation of the landmark positions. These coordinates center the local patches used as input for the prediction model, which then refines these initial projections to determine the precise anatomical locations. For each landmark $\bar{\mathbf{l}}_k^{\text{fit}} \in \mathbb{R}^{1 \times 3}$, a localized patch is extracted from the full facial point cloud $\mathcal{F}_{\text{aligned}}^{\text{og}} $.  Two strategies can be employed to generate these patches. In the first approach, the $K$ nearest points in the aligned facial model $\mathcal{F}_{\text{aligned}}^{\text{og}}$ are selected for each fitted landmark $\bar{\mathbf{l}}_k^{\text{fit}}$, forming a local patch $\mathcal{P}_k \in \mathbb{R}^{K \times 3}$ centered around that landmark. In the second, all points within a fixed radius $D$ from the fitted landmark $\bar{\mathbf{l}}_k^{\text{fit}}$ are selected and then uniformly resampled to $K$ points to ensure a consistent patch size. In both cases, the resulting collection of patches for all landmarks is denoted as $\mathcal{P} = \{ \mathcal{P}_1, \dots, \mathcal{P}_{n_l} \} \in \mathbb{R}^{n_l \times K \times 3}$.  

To ensure consistency, the $K$ points within each patch are further sorted by their Euclidean distance to the origin of the global reference coordinate system \( \mathbf{o} \in \mathbb{R}^{1 \times 3} \) , defined near the nasal region of the face. This structured and repeatable input format improves the model’s ability to capture spatial relationships while reducing training instability that may arise from arbitrary point orderings. The choice of $K$ or $D$ directly affects patch granularity and is further evaluated in the ablation experiments presented in the Results section.

\subsection{Patch-Attention Landmark Network (PAL-Net)}

\begin{figure}[!h]
\centering
\includegraphics[width=1\linewidth]{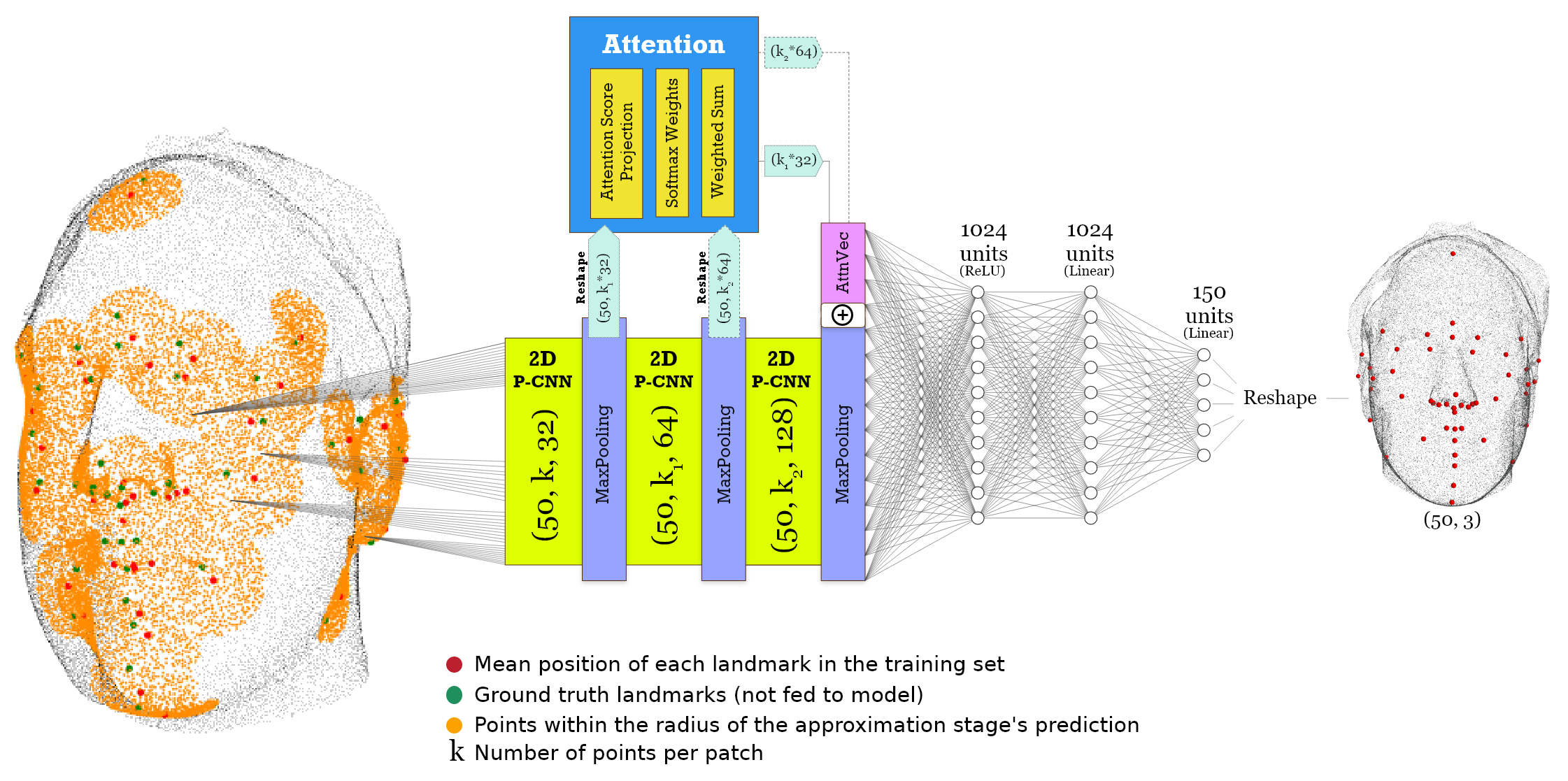}
\caption{\label{fig:models} Architecture overview of PAL-Net for predicting anatomical landmarks on the LAFAS dataset (50 landmarks). On the left, localized facial patches (orange) are extracted around approximated landmark positions (red) based on population-averaged coordinates, with ground truth annotations (green) shown for reference but not used as input. Patches are processed through a series of 2D point-wise CNN blocks with increasing feature depth and max pooling. Attention modules capture global context across all patches. The combined features are passed through fully connected layers to predict the 3D coordinates of 50 landmarks. }
\end{figure}

The proposed Patch-Attention Landmark Network (PAL-Net), demonstrated in Figure \ref{fig:models}, refines 3D facial landmark positions by leveraging both local and global spatial information extracted from structured point cloud patches. The model operates on a structured tensor $\mathcal{X} \in \mathbb{R}^{m \times n \times K \times 3}$, where $m$ is the number of subjects (or batch size), $n$ is the number of landmarks per subject, $K$ is the number of points in each patch, and 3 corresponds to the Cartesian coordinates $(x, y, z)$. To extract features from this spatially ordered representation, PAL-Net employs a series of convolutional layers using point-wise $1 \times 1$ convolutions \citep{hua_pointwise_2018}. Each convolutional block consists of two successive point-wise operations applied independently to each point within a patch:
\[
\mathcal{X}'_{ijkf} = \sigma\left(\sum_{c=1}^{3} W_{cf}^{(l)} \cdot \mathcal{X}_{ijkc} + b_f^{(l)}\right)
\]

where $W^{(l)} \in \mathbb{R}^{3 \times f}$ and $b^{(l)} \in \mathbb{R}^{f}$ denote the learnable parameters of the $l^{\text{th}}$ convolutional layer, $f$ is the number of output feature channels, and $\sigma$ is the activation function, which in our case is a  Rectified Linear Unit (ReLU). The model comprises three sequential convolutional blocks, each consisting of two point-wise $1 \times 1$ convolutional layers with ReLU activations. The first block employs 32 filters, followed by a max pooling operation along the patch dimension ($K$), reducing the number of points by a factor of 5. The second block increases the feature dimensionality to 64 and applies an identical pooling strategy. The third block uses 128 filters and a max pooling operation by a factor of 4. Pooling is selectively applied only along the third axis ($K$), while the second axis (the number of landmarks, $n$) is left unchanged to preserve the structural organization of landmark-wise patches. This strategy ensures that spatial resolution within each patch is progressively reduced, while maintaining the inter-landmark relationships established during the preprocessing stage.

\subsubsection*{Attention Integration}
To incorporate global context and enhance prediction accuracy, PAL-Net integrates an attention mechanism after each convolutional block. Prior to pooling, the feature maps $\mathcal{X}' \in \mathbb{R}^{m \times n \times K \times f}$ are reshaped into a 2D sequence $\mathcal{S} \in \mathbb{R}^{m \times (n \cdot K) \times f}$, concatenating all points across all landmarks for each subject. An attention weight matrix $\mathcal{A} \in \mathbb{R}^{m \times (n \cdot K) \times 1}$ is computed as:
\[
\mathcal{A} = \text{Softmax}(\tanh(\mathcal{S} W + b))
\]
where $W \in \mathbb{R}^{f \times 1}$ and $b \in \mathbb{R}^{1}$ are trainable parameters. The softmax operation ensures that the attention weights across all $n \cdot K$ points sum to 1, producing a probabilistic weighting over the input sequence. The weighted global feature vector is then computed as $\mathcal{G}_i = \sum_{j=1}^{n \cdot K} \mathcal{A}_{ij} \cdot \mathcal{S}_{ij}$, resulting in $\mathcal{G} \in \mathbb{R}^{m \times f}$. This global descriptor captures the most informative spatial features across the full facial geometry and is concatenated with the local features from the final convolutional block, forming a hybrid feature representation $\mathcal{H} \in \mathbb{R}^{m \times n \times h}$, where $h$ is the combined feature dimensionality. The final feature tensor $\mathcal{H}$ is processed by a multilayer perceptron (MLP) composed of three fully connected layers. The first layer maps each feature vector to 1024 dimensions, followed by a ReLU activation and dropout (p=0.1) regularization to introduce nonlinearity and prevent overfitting. The second fully connected layer also has 1024 units and uses a linear activation, allowing the model to adjust the features to the output scale. Finally, a third linear layer projects each feature vector to $\mathbb{R}^{3}$, corresponding to the $(x, y, z)$ coordinates of each anatomical landmark. The output is then reshaped to form the final predicted landmark positions, $\hat{\mathcal{L}} \in \mathbb{R}^{m \times n \times 3}$. 

\subsection{Post-processing}
In practice, the predicted landmark coordinates $\hat{\mathcal{L}} \in \mathbb{R}^{n_l \times 3}$ may not precisely lie on the surface of the 3D facial mesh $\mathcal{F} \in \mathbb{R}^{n_f \times 3}$ due to prediction continuity, mesh sparsity, or corrupted surface regions. To address this, we apply a post-processing step to project each predicted point back onto the mesh. Two strategies are considered: the first projects each predicted point $\hat{\mathbf{l}}_k$ to its nearest surface vertex, i.e., $\hat{\mathbf{l}}_k^{\text{proj}} = \arg\min_{\mathbf{f}_i \in \mathcal{F}} \|\hat{\mathbf{l}}_k - \mathbf{f}_i\|_2$; the second computes the centroid of the $K$ nearest vertices, i.e., $\hat{\mathbf{l}}_k^{\text{cent}} = \frac{1}{K} \sum_{\mathbf{f}_j \in \mathcal{N}_K(\hat{\mathbf{l}}_k)} \mathbf{f}_j$, which provides a more stable estimate in degraded regions of the mesh (e.g., around the ears or hairline) where the surface is incomplete or noisy. In such cases, the ground truth landmark itself may not lie directly on a valid surface vertex. Depending on dataset characteristics and mesh quality, either strategy may be preferable to ensure robust and anatomically plausible outputs.

\section{Experimental analysis}

\subsection{Training Procedure}

The model was trained using a composite loss function designed to balance point-wise accuracy with anatomical consistency. This objective minimizes a localization loss, defined as the mean Euclidean distance between predicted and ground-truth landmarks, combined with a structural loss based on the mean absolute difference of their inter-landmark distances. The final loss was formulated as a weighted sum of the two components, with weights set to $\alpha = 0.6$ and $\beta = 0.4$, prioritizing precise localization while encouraging the preservation of relative distances between landmarks:
\[
\mathcal{L}_{\text{total}} = \alpha \cdot \frac{1}{n} \sum_{k=1}^{n} \left\| \hat{\mathbf{l}}_k - \mathbf{l}_k \right\|_2 + 
\beta \cdot \frac{1}{n^2} \sum_{i=1}^{n} \sum_{j=1}^{n} \left| \left\| \hat{\mathbf{l}}_i - \hat{\mathbf{l}}_j \right\|_2 - \left\| \mathbf{l}_i - \mathbf{l}_j \right\|_2 \right|
\]
$\hat{\mathbf{l}}_k$ and $\mathbf{l}_k$ denote the predicted and ground truth coordinates of the $k$-th landmark, respectively. Model weights were initialized using the Glorot Normal (Xavier normal) method to facilitate efficient gradient propagation during backpropagation. A fixed random seed was used to ensure reproducibility across multiple runs. Training was performed using the Adam optimizer with an initial learning rate of $1 \times 10^{-3}$ and a batch size of 16. To improve convergence and mitigate overfitting, a ReduceLROnPlateau scheduler was employed. The scheduler monitored the validation loss at each epoch and, if no improvement was observed for 8 consecutive epochs, it reduced the learning rate by a factor of 0.5. Additionally, an early stopping mechanism was applied with a patience of 30 epochs, meaning that training would terminate if no further improvement in validation loss was observed for that number of epochs. The model weights corresponding to the epoch with the lowest observed validation loss (typically occurring around epoch 250) were retained and later reloaded to ensure that subsequent evaluations used the optimal parameters. Data preprocessing, patch generation, and landmark approximation were completed prior to training and remained unchanged throughout the training loop.

\subsection{Datasets}
\label{datasets}

\subsubsection{LAFAS Dataset}
The LAFAS dataset served as the primary dataset for this study, and the development of the proposed model was specifically motivated by the need to automate anatomical landmark prediction on this dataset. More specifically, as it contains manually annotated anatomical landmarks with established relevance in real-world biomedical applications, its clinical relevance and standardized acquisition protocol provided an ideal foundation for training and evaluating a robust and generalizable anatomical landmark localization framework. The data were sourced from the LAFAS (Laboratory of Functional Anatomy of the Stomatognathic System of the Dipartimento di Scienze Biomediche per la Salute, Università degli Studi di Milano), which contains 214 3D facial scans of healthy subjects aged 18 to 49 years, both sexes. Excluded from this study were subjects with a previous history of craniofacial traumas, congenital anomalies, or craniofacial surgery. The 3D facial models were acquired using VECTRA M3, a fixed stereophotogrammetric device (Canfield Scientific Inc., Fairfield, NJ, USA), and VECTRA H1, a portable one. These instruments are employed routinely in the LAFAS and have demonstrated comparability and equivalence. In the study by \citep{de2022validation}, the validation of the Vectra 3D imaging system was verified. Each scan includes 50 facial landmarks manually annotated using a standardized protocol developed by \citep{ferrario2003growth}. These landmarks, shown in Figure \ref{fig:landmarks}(a) and listed in table \ref{tab:results}, were manually annotated using a point-and-click interface by the skilled personnel of the LAFAS lab. To improve precision in manual landmark localization, specific landmarks were labeled with eyeliner on the face before acquisition. The 50 anatomical landmarks that cover various facial regions are employed in the LAFAS laboratory, in a variety of biomedical applications, including the assessment of facial asymmetry, morphometric facial characterization, and anthropometric analysis \citep{cappella2023facial, solazzo2025three}. In these analyses, the relative distances between landmarks are typically utilized, rather than their absolute spatial positions. Many clinical and morphological assessments rely on geometric relationships and proportions between landmarks rather than their exact coordinate positions. To align the training objective with these downstream biomedical tasks, the proposed model incorporates a loss component that explicitly penalizes deviations in inter-landmark distances. This design encourages the model not only to accurately localize points but also to preserve the geometric structure of the face, thereby enhancing the clinical relevance and interpretability of the predictions. To assess intra-observer reliability of the LAFAS dataset during the data annotation process, 20 random cases were selected, and the same operator annotated the same set of facial data twice within a two-week interval. This evaluation aimed to determine the consistency of the operator's annotations over time. The results of this reliability analysis are presented in Table \ref{tab:results}, showcasing the mean difference in distance (mean Euclidean distance) between the two rounds of annotations.

\subsubsection{FaceScape Dataset}
\begin{figure}[h]
\centering
\begin{subfigure}[b]{0.35\linewidth}
    \centering
    \includegraphics[width=\linewidth]{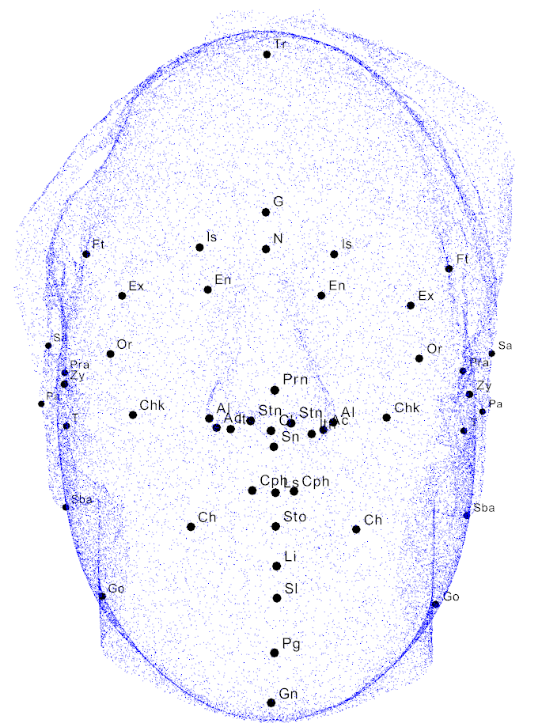}
    \caption{Annotated 3D facial point cloud with anatomical landmarks used in this study, from \citep{sforza2012three}}
    \label{fig:landmarks}
\end{subfigure}
\hspace{2cm} 
\begin{subfigure}[b]{0.35\linewidth}
    \centering
    \includegraphics[width=\linewidth]{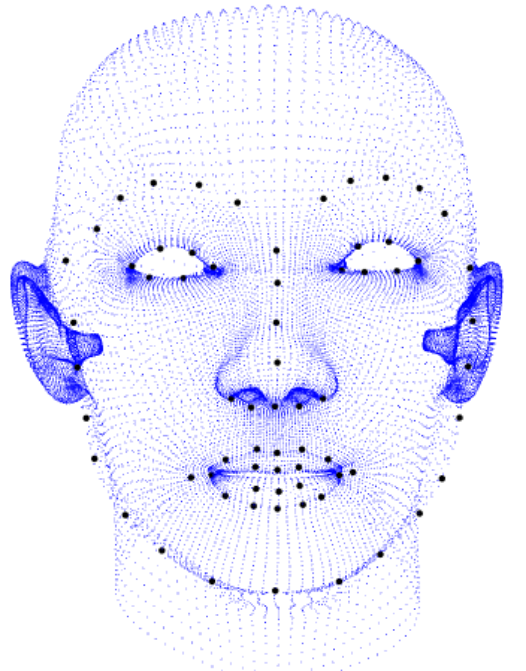}
    \caption{Neutral 3D scan with automatic landmarks used in this study from FaceScape \citep{zhu2023facescape}.}
    \label{fig:second}
\end{subfigure}
\caption{Example 3D facial meshes with annotated landmarks used in this study.}
\label{fig:landmark-comparison}
\end{figure}

The second dataset used in this study was the FaceScape dataset, a publicly available 3D facial dataset introduced by \citet{zhu2023facescape,yang2020facescape}. It contains high-resolution 3D facial scans of over 960 subjects, collected using a multi-camera photogrammetry system. The dataset includes a wide range of facial expressions, head poses, and identity variations, making it a valuable resource for tasks such as 3D face modeling, expression synthesis, and geometry-aware facial analysis. For the purposes of this study, only neutral expression scans were used to ensure consistency with the training data and to provide a clearer evaluation of the model’s generalization to new subjects under similar conditions. Out of the 847 available neutral scans, a subset of 700 subjects was used for training and evaluation. The 3D scans are accompanied by dense mesh correspondence as well as 68 annotated facial landmarks, shown in Figure \ref{fig:landmarks}(b), primarily derived from automatic algorithms rather than manual annotation. While the dataset offers broad subject diversity and expression coverage, the landmark positions are not manually curated and do not follow a clinically validated annotation protocol, unlike the LAFAS dataset, which is tightly linked to biomedical applications and clinical use cases. The FaceScape dataset is primarily designed for general-purpose computer vision, and the annotated landmarks are not routinely used in clinical contexts and may not correspond precisely to anatomical reference points. For this reason, the FaceScape dataset was used solely for exploratory validation and generalization testing, while the model was developed and optimized based on the LAFAS dataset, which better reflects the clinical standards and requirements relevant to the target applications.


\section{Results and Discussion}

\subsection{Comparative analysis with State-of-the-Art Methods}
A comparison with existing methods across the two datasets is reported in Table~\ref{tab:comparison}. All models were trained and evaluated on the same hardware configuration using a Tesla T4 GPU and a batch size of 16 to ensure a fair and consistent benchmarking environment. For the Deep-MVLM model, results correspond to a single training fold, with the best-performing epoch selected from 100 total training epochs. For the LAFAS dataset, results for both 2S-SGCN and the proposed PAL-Net are reported as the mean prediction errors averaged over 5-fold cross-validation. In contrast, for the FaceScape dataset, evaluation was performed on a single fold due to its larger data volume. The increased sample size in FaceScape made full cross-validation computationally intensive and less critical for generalization testing. The hyperparameters of the 2S-SGCN model that was trained on FaceScape were the default configuration reported in the original study, specifically with a model depth of 20 and $k_\text{knn} = 16$. However, for the LAFAS dataset, this configuration did not yield optimal results; instead, a shallower architecture (depth = 4) and a larger local neighborhood ($k_\text{knn} = 32$) provided improved performance, highlighting the importance of dataset-specific tuning. In both datasets, the 2S-SGCN was trained for 500 epochs. These adjustments ensured that each model operated under conditions most suitable to its architecture and the dataset's specific characteristics, enabling a more meaningful performance comparison. All reported results for PAL-Net were obtained using localized facial patches constructed from the 1000 nearest vertices to each approximated landmark in the original facial meshes, trained on around 250 epochs with early stopping enabled. For the LAFAS dataset, the final prediction for each landmark was computed as the centroid of the $K = 10$ closest points in the predicted coordinate space, which helped mitigate local surface noise. In the FaceScape dataset, which contains denser, more complete facial meshes, the final landmark prediction was determined by selecting the surface point closest to the model’s predicted coordinate. In addition to achieving lower mean errors, PAL-Net also demonstrated reduced standard deviation across both point-wise and distance-wise metrics, indicating more consistent predictions and improved robustness. This indicates more consistent predictions across subjects and landmarks, suggesting improved model stability and generalization. Another advantage of PAL-Net lies in its lightweight training characteristics, despite achieving competitive accuracy across both datasets. PAL-Net requires substantially less computational time per training epoch than alternative approaches, and this efficiency is primarily due to the model’s compact architecture and the use of pointwise convolutional operations, which are well-suited for localized patch-based learning. Figure \ref{fig:models_comparison_per_landmark} shows the per-landmark localization error for each landmark of the LAFAS dataset.

\begin{table}[H]
\centering
\small  
\begin{tabular}{llcc}
\toprule
\textbf{Model} & \textbf{Metric} & \makecell{\textbf{LAFAS}\\(214 Subjects) \\ (50 Landmarks)} & \makecell{\textbf{FaceScape}\\(700 Subjects)\\ (68 Landmarks)} \\
\midrule
\multirow{3}{*}{\makecell[l]{MVLM \\ \citep{paulsen_multi-view_2019}}}
& Point-wise (mm)       & $4.691\pm 3.486 $ & $5.254\pm 3.561 $ \\
& Distance-wise (mm)    & $3.870\pm 3.396 $ & $3.975\pm3.399 $ \\
& Train Epoch Time & $40$ min     & $70$ min \\
\midrule
\multirow{3}{*}{\makecell[l]{2S-SGCN \\ \citep{burger20242s}}}
& Point-wise (mm)     & $3.689\pm3.178 $ & $0.709\pm0.944 $ \\
& Distance-wise (mm)  & $2.973\pm3.069 $ & $0.613\pm0.778 $ \\
& Train Epoch Time    & $41.4$ sec  & $81$ sec \\
\midrule
\multirow{3}{*}{\makecell[l]{\textbf{PAL-Net} \\ (Ours)}}
& Point-wise (mm)     & $\bm{3.686\pm2.306}$ & $\bm{0.410\pm0.634}$ \\
& Distance-wise (mm)  & $\bm{2.822\pm2.326}$ & $\bm{0.380\pm0.536}$ \\
& Train Epoch Time    & \bm{$3.32$} sec  & \bm{$7.51$} sec \\
\bottomrule
\end{tabular}
\caption{Comparison of point-wise and distance-wise localization errors (in mm) for different methods on LAFAS and FaceScape datasets on the evaluation set corresponding to 20\% of the total number of cases in each dataset. PAL-Net achieved the best accuracy with a significantly lower GPU memory requirement (2.5 GiB vs. 9–11 GiB).\label{tab:comparison} }
\end{table}

To evaluate the computational efficiency of PAL-Net in comparison with existing approaches, we measured the average runtime per subject on the LAFAS dataset, broken down into three stages: Input Preparation, Model Inference, and Final Prediction, as reported in Table~\ref {tab:runtime_comparison}. These measurements allow a detailed comparison of the end-to-end runtime characteristics of each method under realistic conditions. All evaluations were performed using an NVIDIA Tesla T4 GPU, and I/O-related delays (e.g., file imports) were explicitly excluded to focus on model-relevant computations. Inference timing was performed with a batch size of 1, and all other times are averaged per subject to ensure subject-level precision. For PAL-Net, input preparation includes generating localized point cloud patches and sorting the points; inference covers the model’s forward pass; and the final prediction step projects the outputs onto the mesh surface using nearest-neighbor or $K$-point averaging. For MVLM, input preparation includes rendering RGB views from 3D models; inference corresponds to heatmap-based landmark prediction; and final prediction involves back-projecting 2D predictions onto the 3D surface. For 2S-SGCN, input preparation includes graph construction; inference refers to the forward pass; and final prediction computes the MSE-over-mesh. PAL-Net achieves the lowest inference time, while its total runtime remains competitive despite additional patch creation and post-processing.

\begin{table}[H]
\centering
\setlength{\tabcolsep}{4pt} 
\begin{tabular}{lcccc}
\toprule
\textbf{Model} & \makecell{\textbf{Input}\\\textbf{Preparation}} & \makecell{\textbf{Model}\\\textbf{Inference}} & \makecell{\textbf{Final}\\\textbf{Prediction}} & \makecell{\textbf{Total}\\\textbf{Time}} \\
\midrule

\makecell[l]{Manual Annotation }  & – & – & – & 6-7 min \\ 
\makecell[l]{MVLM  \citep{paulsen_multi-view_2019}}     & 0.431 sec & 11.552 sec & 5.101 sec         & 12.084 sec \\
\makecell[l]{2S-SGCN \citep{burger20242s}}             & \textbf{0.062 sec} & 0.102 sec & \textbf{0.042 sec} & \textbf{0.206 sec} \\
\makecell[l]{\textbf{PAL-Net}  (Ours)}                  & 0.207 sec & \textbf{0.005 sec} & 0.092 sec & 0.304 sec \\
\bottomrule
\end{tabular}
\caption{Average runtime per subject (in seconds) for each model, broken down into input preparation, inference, and post-processing (final prediction). Manual annotation time by an expert is included for comparison. \label{tab:runtime_comparison}}
\end{table}

\subsection{Anatomical Analysis}

In the following section, we present a comprehensive evaluation of PAL-Net on the LAFAS dataset based on two complementary analyses: point-wise and distance-wise, using results obtained through a 5-fold cross-validation framework. This evaluation strategy ensures that the reported performance reflects the model’s ability to generalize across different subsets of the data while reducing bias introduced by a single train-test split.

\subsubsection{Point-Wise Analysis}
Euclidean distances between predicted and ground-truth landmarks were computed for each validation sample. Averaging these across folds yielded mean errors and standard deviations for all 50 landmarks. As detailed in Table \ref{tab:results}, these metrics quantify the model's point-wise accuracy and generalization capability. Overall, PAL-Net achieved a mean localization error of 3.686 ± 2.306 mm, demonstrating robust and consistent performance across most landmarks. As expected, midline landmarks such as Subnasale (Sn), Stomion (Sto), and Nasion (N) exhibit relatively low errors, often below 2.5 mm. These landmarks tend to lie in well-defined, geometrically stable facial regions, which makes them easier to learn and to predict reliably. Similarly, paired landmarks on geometrically symmetrical areas such as Orbitales (Or), Cheek Points (Chk), and Labiale Superius/Inferius (Ls, Li) show closely matching errors between their left and right counterparts, indicating that PAL-Net maintains spatial symmetry and is not biased toward one side of the face. Landmarks located around the ear region, such as Gonion (Go), Tragion (T), Postaurale (Pa), and Superaurale (Sa), along with the Trichion (Tr), consistently exhibited the highest localization errors, frequently exceeding 5 mm and reaching over 7 mm in some cases. This elevated error is primarily attributed to corruption in the input mesh in these regions. For the majority of subjects, the 3D surface data around the ears and hairline were either incomplete, noisy, or poorly defined, largely due to occlusion and scattering introduced by hair. In the case of Trichion, the landmark was often located above the dense hairline, where the facial mesh abruptly degrades or ends, resulting in landmark positions that were not even on the mesh's valid surface. As a corrective measure during postprocessing, the mean position of the 10 closest vertices to the predicted landmark coordinate was assigned as the final predicted landmark to mitigate the effects of missing or corrupted geometry. Despite these efforts, these regions remain particularly challenging, and the high variability in surface quality limits the model’s ability to consistently localize landmarks with high precision. The Gonion (Go) landmark consistently ranks among the highest localization errors in 3D facial landmarking studies and was also among the least accurately predicted by PAL-Net. In the study by \citet{aldridge2005precision}, which assessed the precision and measurement error of 3D soft tissue landmarks using photogrammetric systems, Gonion showed the highest localization error among all landmarks, with mean deviations up to 4.10 mm and a standard deviation of 1.64 mm. Moreover, inter-landmark distances involving Gonion were among those with the highest digitization errors, exceeding 5\% of the total variance, indicating operator difficulty with consistent placement. This reduced accuracy is primarily due to the limited visibility of the landmark and the variable soft-tissue coverage in the mandibular angle region, making both manual and automated identification more challenging \citep{staller2022precision, nord20153dmd}.  Interestingly, landmarks such as the Pogonion (Pg) and Crista Philtri (Cph) exhibit low mean errors (~2.5–2.8 mm) despite being near highly curved regions, suggesting that the model generalizes well in structurally challenging areas when sufficient training data coverage is available. The standard deviation values across folds are generally low (0.297 mm on average), indicating stable model behavior and low sensitivity to fold-specific variability. This reflects both the anatomical consistency of the LAFAS dataset and the architectural regularization effect introduced by the attention modules in PAL-Net. Taken together, the results indicate that PAL-Net performs best in well-defined and midline regions, with consistent generalization across folds, while challenges remain in areas with surface noise or structural variability. These findings align with known limitations in 3D facial landmark localization and further emphasize the importance of geometric quality in input data for precise prediction.

To contextualize the performance of PAL-Net, it is important to compare the model’s localization accuracy to known measures of intra-observer and inter-observer variability. In our case, the tight intra-observer variability across the dataset averages around $0.798 \text{ mm}$. PAL-Net achieves an overall mean localization error of $3.686 \pm 2.306 \text{ mm}$ across the 50 landmarks. While this error margin exceeds the precision baseline set by a single expert, it is critical to acknowledge that this performance is often constrained by data scarcity inherent to specialized medical datasets. Furthermore, the performance is not uniform, a substantial number of the 50 landmarks achieve significantly lower error, approaching human intra-observer consistency, while the overall mean is elevated due to higher variability in a smaller number of anatomically complex landmarks. Despite the precision gap with the gold standard, the $3.686 \text{ mm}$ error remains within a clinically reasonable range for many applications. PAL-Net's value lies not in surpassing human precision, but in providing a consistent, standardized, and highly automated alternative.


\begin{longtable}{lc|c}
\toprule
\textbf{Landmark} & 
\multicolumn{1}{c}{\makecell{\textbf{Intra-}\\\textbf{observer}\\\textbf{Reliability (mm)}}} & 
\multicolumn{1}{c}{\makecell{\textbf{Localization}\\\textbf{Accuracy (mm)}}} \\
\midrule
\endfirsthead
\multicolumn{3}{c}%
{{\bfseries \tablename\ \thetable{} -- continued from previous page}} \\
\toprule
\textbf{Landmark} & 
\multicolumn{1}{c}{\makecell{\textbf{Intra-}\\\textbf{observer}\\\textbf{Reliability (mm)}}} & 
\multicolumn{1}{c}{\makecell{\textbf{Localization}\\\textbf{Accuracy (mm)}}} \\
\midrule
\endhead
\midrule \multicolumn{3}{r}{{Continued on next page}} \\
\endfoot
\bottomrule
\endlastfoot
Trichion (Tr)                               & 0.4289 & 6.413 ± 4.614 \\
Glabella (G)                                & 0.7238 & 2.672 ± 1.862 \\
Nasion (N)                                  & 0.7766 & 2.240 ± 1.498 \\
Pronasale (Prn)                             & 0.4231 & 2.096 ± 1.194 \\
Columella (C)                               & 0.4932 & 1.824 ± 1.086 \\
Subnasale (Sn)                              & 0.4711 & 1.653 ± 0.967 \\
Labiale Superius (Ls)                       & 0.5104 & 2.035 ± 1.191 \\
Stomion (Sto)                               & 0.3407 & 1.640 ± 0.964 \\
Labiale Inferius (Li)                       & 0.5875 & 1.995 ± 1.251 \\
Sublabiale (Sl)                             & 0.9986 & 2.404 ± 1.558 \\
Pogonion (Pg)                               & 0.9200 & 2.750 ± 1.669 \\
Gnathion (Gn)                               & 0.8944 & 4.218 ± 3.279 \\
Tragion  Right (T)                          & 0.6925 & 4.277 ± 2.161 \\
Preaurale  Right (Pra)                      & 1.5077 & 5.437 ± 3.201 \\
Superaurale  Right (Sa)                     & 1.3221 & 6.505 ± 3.965 \\
Postaurale  Right (Pa)                      & 1.5267 & 6.940 ± 4.468 \\
Subaurale  Right (Sba)                      & 1.2115 & 5.130 ± 2.903 \\
Frontotemporale  Right (Ft)                 & 0.5389 & 4.823 ± 2.586 \\
Zygion  Right (Zy)                          & 0.3994 & 5.345 ± 3.082 \\
Gonion  Right (Go)                          & 0.3803 & 7.169 ± 3.916 \\
Orbitale Superius  Right (Os)               & 0.4817 & 3.407 ± 2.053 \\
Exocanthion  Right (Ex)                     & 0.8051 & 2.806 ± 1.528 \\
Orbitale  Right (Or)                        & 0.3592 & 5.613 ± 3.719 \\
Endocanthion  Right (En)                    & 0.4633 & 2.206 ± 1.792 \\
Malare Cheek  Right (Chk)                   & 1.5841 & 3.899 ± 2.314 \\
Alar crest  Right (Ac)                      & 0.5654 & 2.202 ± 1.319 \\
Alare  Right (Al)                           & 0.6430 & 1.960 ± 1.263 \\
\makecell[l]{Inferior terminal point \\\hspace{3mm}of the nostril axis  Right (Itn)} & 0.8018 & 1.900 ± 1.064 \\
\makecell[l]{Superior terminal point \\\hspace{3mm}of the nostril axis  Right (Stn)} & 0.5100 & 1.866 ± 1.057 \\
Crista Philtri  Right (Cph)                 & 0.5949 & 2.134 ± 1.097 \\
Cheilion  Right (Ch)                        & 0.4834 & 2.586 ± 1.496 \\
Tragion  Left (T)                           & 0.7178 & 3.688 ± 2.011 \\
Preaurale  Left (Pra)                       & 2.0102 & 5.059 ± 3.274 \\
Superaurale  Left (Sa)                      & 1.5333 & 6.041 ± 3.902 \\
Postaurale  Left (Pa)                       & 1.4705 & 6.709 ± 3.913 \\
Subaurale  Left (Sba)                       & 1.7314 & 4.889 ± 2.813 \\
Frontotemporale  Left (Ft)                  & 0.4314 & 4.629 ± 2.493 \\
Zygion  Left (Zy)                           & 0.3389 & 5.619 ± 4.213 \\
Gonion  Left (Go)                           & 0.4850 & 7.309 ± 5.048 \\
Orbitale Superius  Left (Os)                & 0.3231 & 3.576 ± 2.123 \\
Exocanthion  Left (Ex)                      & 0.8059 & 3.226 ± 2.862 \\
Orbitale  Left (Or)                         & 0.4385 & 5.779 ± 3.751 \\
Endocanthion  Left (En)                     & 0.4930 & 2.491 ± 2.493 \\
Malare Cheek  Left (Chk)                    & 2.1667 & 4.389 ± 2.864 \\
Alar crest  Left (Ac)                       & 1.0431 & 2.204 ± 1.323 \\
Alare  Left (Al)                            & 1.1591 & 2.068 ± 1.158 \\
\makecell[l]{Inferior terminal point \\\hspace{3mm} of the nostril axis  Left (Itn)} & 0.7360 & 1.911 ± 1.121 \\
\makecell[l]{Superior terminal point \\\hspace{3mm} of the nostril axis  Left (Stn)} & 0.5504 & 1.977 ± 1.094 \\
Crista Philtri  Left (Cph)                  & 0.5784 & 2.105 ± 1.200 \\
Cheilion  Left (Ch)                         & 0.4944 & 2.489 ± 1.534 \\\\
\textbf{Mean}                      & \textbf{0.798 mm}& \textbf{3.686 ± 2.306 mm} \\
\bottomrule
\caption{\label{tab:results}Localization accuracy of anatomical landmarks, averaged over the validation sets of each fold, with  standard deviation of the localization errors computed independently for each fold, then averaged across all folds.} \\

\end{longtable}

\subsubsection{Distance-Wise Analysis}

\begin{figure}[!h]
\centering
\includegraphics[width=0.8\linewidth]{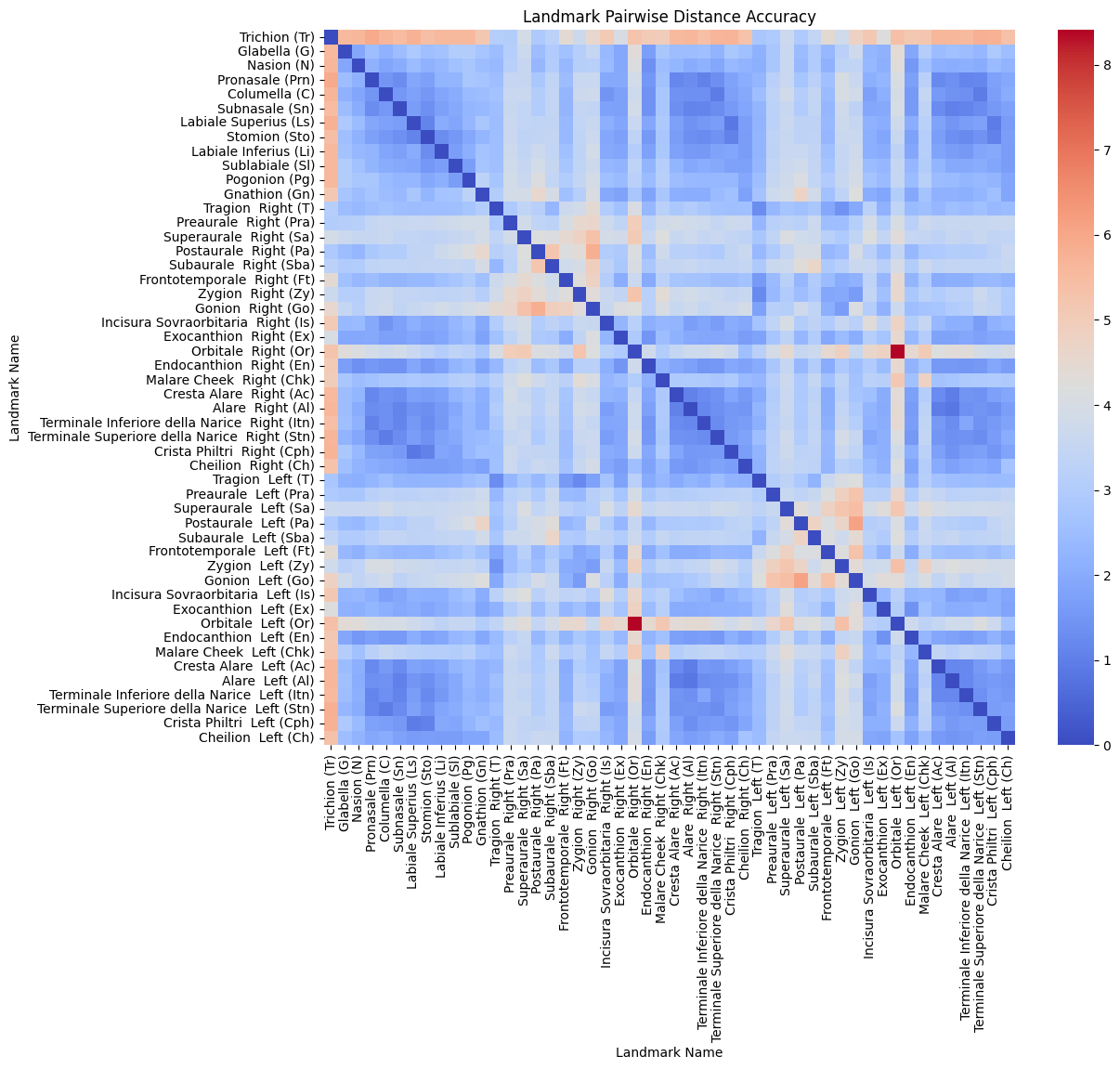}
\caption{\label{fig:distance_matrix} Average distance-wise error matrix between ground truth and predicted landmarks averaged over the 5-fold cross validation. Each entry represents the absolute difference in pairwise distances for a specific landmark pair, averaged over the test set. The mean of the matrix is 2.822mm, indicating the overall distance-wise error.}
\end{figure}

To assess preservation of spatial relationships between landmarks, we performed a pairwise distance-variability analysis using 5-fold cross-validation. This involved computing inter-landmark distances and comparing their differences across the predicted and ground-truth datasets for each fold. The matrix shown in Figure \ref{fig:distance_matrix} is derived by calculating the average distance-wise error for each pair of landmarks across all test cases in the dataset. For each pair of landmarks, such as A and B, the Euclidean distance is computed independently for the ground-truth and predicted landmark positions. The absolute difference between these distances is then calculated for all test cases in the dataset, capturing how accurately the predicted landmarks maintain the spatial relationship between each pair. Finally, the differences are averaged across the entire set, yielding a symmetrical error matrix in which each entry represents the average absolute error for a specific pair of landmarks. The mean value of the matrix is 2.822mm, representing the overall distance-wise error between the predicted and ground-truth landmarks. From the figure, we observe that the matrix shows areas where the model struggles to replicate ground-truth distances. Specifically, errors in the first row and column indicate that the model has difficulty preserving the spatial consistency of the Trichion (Tr) landmark relative to all other landmarks. This suggests that the model's predictions for Trichion are less reliable; similarly, the paired landmarks of Orbitale showed the largest error (the hottest point in the matrix). Conversely, most other landmark pairs exhibit smaller errors, showing that the model better maintains their spatial relationships.

Since this study focuses on anatomical landmarks for medical anthropometric analysis, not all linear distances on the face hold equal significance. Some facial linear distances are more distinct and clinically relevant than others \citep{pucciarelli2017face}. To ensure meaningful analysis, specific anatomical linear distances and angles were selected for further evaluation. These measurements were then compared with the intra-observer variability to provide a comprehensive assessment of the predicted facial landmarks and their accuracy. Table \ref{tab:Linear_distances} demonstrates the intra-observer variability of anatomical distances, predicted landmarks' distances, and the ratio between the predicted landmarks' distances and the average distances of the landmarks. The ratio was calculated to provide insight into the significance of the error, depending on the distance. This approach allows evaluation of the magnitude of the error relative to the actual anatomical distances being measured, providing a clearer understanding of the model's performance. 

Despite the 2 mm benchmark being widely accepted as a clinically relevant threshold for facial landmark localization accuracy \citep{dindarouglu2016accuracy, othman2020validity, weinberg2006anthropometric}, not all linear distances predicted by our model fall within this margin. Anatomical distances vary greatly in scale, ranging from short inter-landmark distances such as (Cph)R—(Cph)L (~11.8 mm, average distance) to broader measurements such as (Zy)R—(Zy)L (~131.7 mm, average distance). In such cases, reporting error in absolute millimeters may not fully capture the model’s performance. Therefore, we normalized the error by the average ground-truth distances to provide relative insight. The majority of the predicted distances show errors well below 5\% of the corresponding anatomical length, with a mean relative error of 4\%. This ratio is informative when the absolute errors exceed 2 mm, while the proportional error remains within an acceptable range for clinical and anthropometric applications. When considered relative to landmark spacing and clinical practice tolerance, the results strongly support the feasibility of the proposed method for accurate anatomical landmark prediction. These results support the robustness of the model across both short and long facial spans, with predictive accuracy aligning well with clinical tolerances and intra-observer variability.

\begin{table}[H]
\centering
\small
\begin{tabular}{lccc}
\toprule
\textbf{Linear Distances} & 
\multicolumn{1}{c}{\makecell{\textbf{Intra} \\\textbf{Variability}\\\textbf{(mm)}}} & 
\multicolumn{1}{c}{\makecell{\textbf{Error} \\\textbf{Distances}\\\textbf{(mm)}}} & 
\multicolumn{1}{c}{\makecell{\textbf{Error Distances} \\ \textbf{Avg. Distance (\%)}}}
 \\
\midrule
\textbf{Frontal Distances} && \\
(Tr) — (N)  & 0.588 & 5.615 & 8.98\% \\
(N) — (Pg)  & 0.890 & 2.726 & 2.64\% \\
(N) — (Sn)  & 0.444 & 1.990 & 3.76\% \\
(Sn) — (Pg) & 0.840 & 2.290 & 4.43\% \\\\

\textbf{Horizontal Plane} && \\
(Ex)R — (Ex)L   & 0.939 & 2.747 & 3.17\% \\
(Zy)R — (Zy)L   & 0.081 & 1.942 & 1.47\% \\
(T)R — (T)L     & 0.192 & 1.250 & 0.91\% \\
(Ch)R — (Ch)L   & 0.652 & 2.709 & 5.68\% \\
(Cph)R — (Cph)L & 0.738 & 1.412 & 11.94\% \\
(Go)R — (Go)L   & 0.328 & 4.101 & 3.69\% \\\\

\textbf{Sagittal Plane (Right)} && \\
(T)R — (N)      & 0.459 & 2.321 & 2.02\% \\
(T)R — (Sn)     & 0.415 & 2.454 & 2.07\% \\
(T)R — (Pg)     & 0.544 & 2.703 & 2.04\% \\
(Pg) — (Go)R    & 0.487 & 3.836 & 4.01\% \\
(T)R — (Go)R    & 0.542 & 4.371 & 8.06\% \\\\

\textbf{Sagittal Plane (Left)} && \\
(T)L — (N)      & 0.428 & 2.058 & 1.81\% \\
(T)L — (Sn)     & 0.510 & 2.216 & 1.88\% \\
(T)L — (Pg)     & 0.700 & 2.509 & 1.91\% \\
(Pg) — (Go)L    & 0.384 & 3.975 & 4.18\% \\
(T)L — (Go)L    & 0.503 & 4.172 & 7.70\% \\\\
\textbf{Mean} & \textbf{0.533 mm}& \textbf{2.870 mm}& \textbf{4.11\%} \\
\bottomrule
\end{tabular}
\caption{\label{tab:Linear_distances}  Mean absolute errors of selected linear anatomical distances, averaged over the validation sets of each fold. The table reports intra-observer variability, predicted distance errors, and the normalized error expressed as a percentage of the corresponding anatomical distance. }
\end{table}

In addition to evaluating linear anatomical distances, we quantified how accurately PAL-Net reproduces clinically relevant anatomical angles defined by \citep{de2010accuracy, gibelli_are_2018} (see Table \ref{tab:angular_accuracy}). This table compares intra-observer variability and prediction accuracy for the anatomical angles, providing further insight into the accuracy and reliability of the predicted landmarks. Across nine standard angles, the mean absolute error was 1.28°, corresponding to just a 2.1 \% deviation from average anatomical values. Most midline and transverse angles, for example, (T)R-(Prn)-(T)L and (N)-(Prn)-(Pg) , showed errors below 1°, reflecting consistent performance in regions critical to facial symmetry. Larger deviations appeared in measures involving more complex mandibular contours, notably (Go)R–(Pg)–(Go)L (2.86°, 4\% relative error) and (Sn)–(N)–(Prn) (1.44°, 8\% relative error), likely due to inherent anatomical variability at these landmarks. Overall, PAL-Net’s angular accuracy closely approaches expert manual annotations and remains within accepted clinical tolerances for the majority of key anthropometric parameters, further supporting its use for automated, high-throughput 3D facial landmarking in both research and clinical workflows.

\begin{table}[H]
\centering
\begin{tabular}{lc|cc}
\toprule
\textbf{Angular Distances} & 
\multicolumn{1}{c}{\makecell{\textbf{Intra}\\\textbf{Variability}\\\textbf{($^{\circ}$)}}} & 
\multicolumn{1}{c}{\makecell{\textbf{Error}\\\textbf{Angles}\\\textbf{($^{\circ}$)}}} & 
\multicolumn{1}{c}{\makecell{\textbf{Error angle/}\\\textbf{Avg. Angle} \\\textbf{(\%)}}} \\
\midrule

(T)R - (N) - (T)L       & 0.095 & 1.190     &1.33\% \\
(T)R - (Prn) - (T)L     & 0.032 & 0.656     &1.00\% \\
(T)R - (Pg) - (T)L      & 0.345 & 0.446     &0.66\% \\
(Go)R - (Pg) - (Go)L    & 0.176 & 2.863     &4.33\% \\
(N) - (Sn) - (Pg)       & 1.435 & 1.260     &1.00\% \\
(N) - (Prn) - (Pg)      & 0.682 & 0.880     &0.66\% \\
(Sn) - (N) - (Prn)      & 0.228 & 1.440     &8.00\% \\
(T)R - (Go)R - (Pg)     & 1.618 & 1.916     &1.66\% \\
(T)L - (Go)L - (Pg)     & 0.676 & 0.910     &1.00\% \\\\

\textbf{Mean} & \textbf{0.587$^{\circ}$}& \textbf{1.284$^{\circ}$}& \textbf{2.18\%} \\
\bottomrule
\end{tabular}
\caption{\label{tab:angular_accuracy} Mean absolute errors of selected anatomical angles, averaged over the validation sets of each fold. The table includes intra-observer variability, predicted angular errors, and the normalized error expressed as a percentage of the corresponding average anatomical angle. }
\end{table}

\subsection{Ablation Study}

To assess the contributions of different aspects of the PAL-Net model, such as the attention mechanism and model depth, we conducted an ablation study involving alternative model configurations. Each variant was trained and evaluated using 5-fold cross-validation on the LAFAS dataset, maintaining identical preprocessing and training procedures for consistency. The first variant entirely removed the attention modules to evaluate their impact on landmark localization accuracy. The second employed a Top-K attention mechanism, selecting the $k$ most informative points per attention layer to assess computational complexity and test whether selective context aggregation could improve performance. The third variant reduced the network depth by using two convolutional blocks instead of three, investigating the effect of a shallower model capacity on predictive accuracy. Table~\ref{tab:ablation} summarizes the mean localization errors across folds for each configuration. None of the ablated models outperformed the baseline PAL-Net with attention and three convolutional blocks. Specifically, the absence of attention led to a degradation in performance, highlighting the importance of global context modeling. Similarly, the Top-K attention variant did not yield improvements, resulting in higher complexity. The reduced-depth model also showed lower accuracy, indicating that sufficient network capacity is essential for capturing the complex spatial patterns required for precise 3D facial landmark localization. These results confirm that the original PAL-Net architecture balances model complexity and context integration effectively for this task.

\begin{table}[ht]
\centering
\label{tab:ablation}
\begin{tabular}{lcc}
\toprule
\textbf{Model Variant / Preprocessing} & 
\multicolumn{1}{c}{\makecell{\textbf{Localization} \\ \textbf{Error (mm)}}} & 
\multicolumn{1}{c}{\makecell{\textbf{Distance} \\ \textbf{Error (mm)}}}\\
\midrule
\multicolumn{3}{l}{\textbf{Preprocessing Variants (Applied to Baseline Architecture)}} \\
\midrule
Fixed-size patches (500 points)   & $3.731$ & $2.866$ \\
Fixed-size patches (1500 points)  & $3.800$ & $2.878$ \\
Radius-based patches (10 mm)      & $3.765$ & $2.889$ \\
Radius-based patches (15 mm)      & $3.758$ & $2.896$ \\
Radius-based patches (20 mm)      & $3.747$ & $2.886$ \\
PAL-Net (coarse mesh)             & $3.985$ & $3.041$ \\
PAL-Net (no spatial ordering)     & $3.773$ & $2.872$ \\
PAL-Net (with RGB texture)        & $4.293$ & $3.172$ \\\\
\midrule
\multicolumn{3}{l}{\textbf{Model Architecture Variants}} \\
\midrule
\textbf{Baseline PAL-Net (1000 points)} & $\bm{3.686}$& $\bm{2.822}$\\
Without attention modules       & $3.713$& $2.851$\\
Top-K attention ($k=10$)        & $3.824$& $2.897$\\
Reduced depth (2 conv layers)   & $3.909$& $2.964$\\
\bottomrule
\end{tabular}
\caption{ \label{tab:ablation} Ablation study results: mean localization and distance errors (in mm) averaged across 5-fold cross-validation.}
\vspace{0.5em}
\raggedright
\end{table}

Further analysis was performed to investigate the influence of preprocessing parameters on PAL-Net’s landmark localization performance. The baseline model was trained using two distinct patch extraction strategies to evaluate its sensitivity to local neighborhood definitions and point cloud density. In the first approach, patches were extracted by selecting a fixed number of surface points closest to each approximated landmark, with patch sizes of $k = $  500, 1000, and 1500 points. This strategy ensured consistent input dimensionality across samples, while allowing control over local detail granularity. In the second approach, patches were generated by including all surface points within fixed Euclidean radii of $D = $  10 mm, 15 mm, and 20 mm from the approximated landmark positions. This method simulates scenarios involving sparser or variable-resolution meshes, such as those commonly encountered in low-cost or mobile acquisition systems.  
The results demonstrate that the baseline PAL-Net model using 1000-point, fixed-size patches achieves the best overall performance, with localization and distance errors of $3.686$\,mm and $2.822$\,mm, respectively. Reducing the number of points to 500 slightly increases the error, while using 1500 points does not yield further improvement, suggesting a saturation point around 1000 points. Radius-based patching with radii of 10-20\,mm results in comparable performance ($3.74$-$3.76$\,mm localization error), indicating that increasing spatial extent does not provide significant benefit. These results emphasize that careful tuning of patch size, both in terms of point count and spatial coverage, is important to balance local detail capture and noise suppression.
To evaluate PAL-Net's robustness to lower-resolution inputs, we tested its performance on coarsely sampled 3D facial meshes. Instead of constructing patches from the original high-resolution meshes, we downsampled each mesh to 10,000 points using uniform surface sampling. Local patches were then extracted from this coarse representation, selecting only the $K = 100$ nearest points around each approximated landmark, compared to $K = 1000$ in the baseline configuration. This setting simulates applications where acquisition devices produce sparse point clouds (e.g., mobile or low-cost systems). Despite the reduced point density and smaller patch size, PAL-Net retained performance, demonstrating its adaptability to varying mesh resolutions. 
To investigate the role of spatial ordering within each patch, we trained a variant of PAL-Net that does not enforce consistent point ordering. While point-wise convolutional layers are permutation-invariant, the fully connected layers at the end of the network aggregate information across all points, introducing sensitivity to input ordering. Without consistent ordering, the model receives arbitrary point arrangements for each patch, which hinders its ability to capture stable spatial patterns. Removing spatial ordering led to a measurable drop in performance, confirming that consistent point arrangement facilitates more effective feature aggregation and improves localization accuracy. The corresponding results are reported in Table \ref{tab:ablation}.
To address the relevance of texture information, we evaluated a variant of PAL-Net that included RGB color values appended to the $(x, y, z)$ coordinates of each point. As shown in Table \ref{tab:ablation}, incorporating texture did not improve performance and actually led to a slight degradation in localization accuracy. This suggests that, for anatomical landmarking, geometric surface features are the primary discriminators, while texture data may introduce noise from lighting variations or skin tone that obscures the underlying morphological structure.

\begin{figure}[!h]
\centering
\includegraphics[width=1\linewidth]{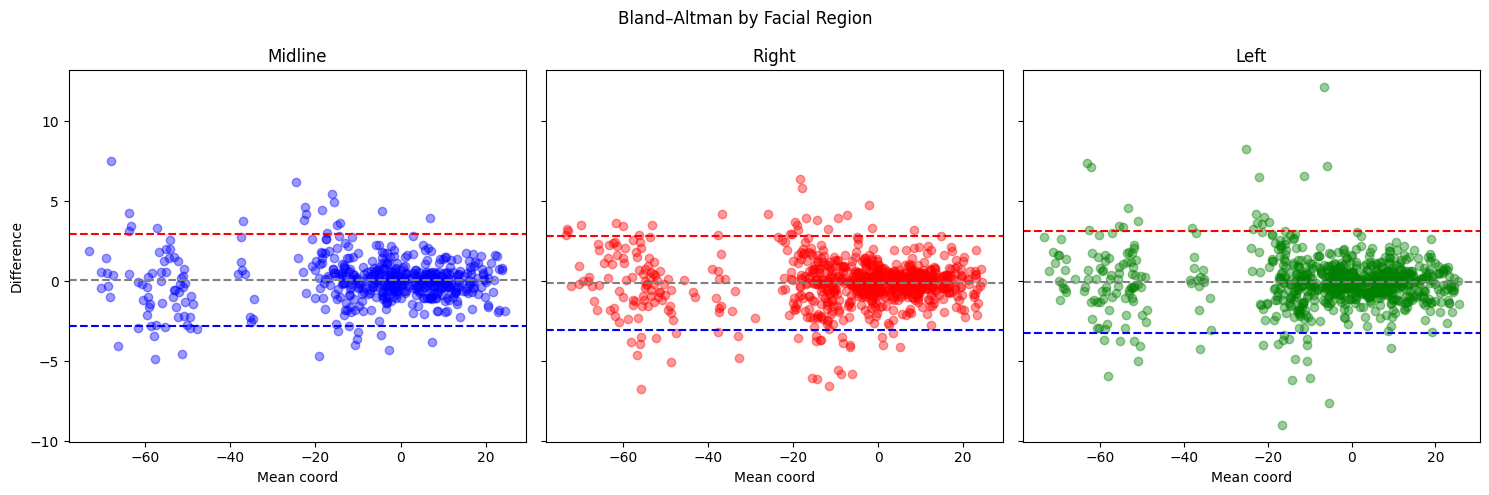}
\caption{\label{fig:baldman} Bland–Altman plots showing prediction errors by facial region (midline, right, left). Each plot shows the difference between predicted and ground-truth coordinates relative to their mean. The dashed lines represent the mean difference (gray) and the 95\% limits of agreement (red and blue). The results indicate no systematic bias across regions, with most predictions falling within acceptable error bounds.}
\end{figure}

To evaluate whether prediction accuracy varied across facial regions, a Bland–Altman analysis was performed separately for midline, left, and right landmarks (Figure \ref{fig:baldman}). In all three regions, the mean difference between predicted and ground-truth coordinates was close to zero, indicating no systematic bias. Most differences fell within the 95\% limits of agreement, with no strong trend suggesting under- or over-prediction in any specific region. The spread of differences was slightly wider for lateral landmarks, particularly on the left side, which may reflect localized variability in mesh quality or anatomical asymmetry. Overall, the plots suggest consistent model behavior across facial regions, with no evidence of directional error or regional bias.

\subsubsection*{Exclusion of Peripheral Landmarks}

Certain landmarks located around the ears, namely \textit{Preaurale (Pra), Superaurale (Sa), Postaurale (Pa), and Subaurale (Sba)} on both the left and right sides, consistently exhibited higher localization errors across subjects. These points are often affected by mesh corruption caused by hair, leading to incomplete or noisy surface geometry. In several cases, the ground-truth landmark did not lie on the surface of any mesh point, making accurate prediction and evaluation inherently ambiguous. To better assess PAL-Net's performance and that of other methods in clinically relevant, geometrically stable regions, we excluded the eight peripheral landmarks and recomputed the localization and distance-based errors. Table~\ref{tab:comparison_face_only} presents the updated evaluation, showing improved performance when focusing solely on core facial landmarks. This adjustment provides a more representative metric of the model's effectiveness for central facial analysis tasks.

\begin{table}[H]
\centering
\begin{tabular}{lccc}
\toprule
\textbf{Model} & 
\multicolumn{1}{c}{\makecell{\textbf{Point-wise} \\ \textbf{Error (mm)}}} &
\multicolumn{1}{c}{\makecell{\textbf{Distance-wise} \\ \textbf{Error (mm)}}}\\
\midrule
MVLM~\citep{paulsen_multi-view_2019}     & $4.342 $ & $3.742 $\\
2S-SGCN~\citep{burger20242s}             & $\bm{3.196} $ & $3.076 $ \\
PAL-Net (Ours)                          & $3.276$ & $\bm{2.570}$ \\
\bottomrule
\end{tabular}
\caption{Comparison of point-wise and distance-wise localization errors (in mm) for different methods on the LAFAS dataset after excluding peripheral ear-region landmarks (8 out of 50 landmarks). Metrics reflect performance over the remaining 42 facial landmarks. \label{tab:comparison_face_only}}
\end{table}

On the reduced set of 42 facial landmarks (excluding the 8 ear-region points), 2S-SGCN has slightly lower point-wise accuracy than PAL-Net ($3.196$,mm vs. $3.276$,mm). However, PAL-Net maintains the best performance in terms of distance-wise accuracy ($2.570$,mm), suggesting that its predictions better preserve anatomical structure. The distance-wise error of 2S-SGCN slightly increases compared to the full landmark set, indicating reduced spatial coherence when peripheral landmarks are excluded. This further underscores PAL-Net’s strength in maintaining geometric consistency across core facial structures.

\subsubsection*{Discussion on Model Interpretability}

Unlike traditional geometric methods that rely on explicit, hand-crafted features, deep learning models are often criticized for their lack of transparency. While we do not visualize internal feature maps, our experimental analysis provides critical insight into how PAL-Net arrives at its predictions by isolating the contribution of different spatial contexts.
First, the ablation study (Table \ref{tab:ablation}) explains the role of global context. The performance drop observed when removing the attention mechanism confirms that local geometric features alone are insufficient for high-accuracy landmarking. The model explicitly relies on the attention-weighted aggregation of features from distant patches to resolve ambiguities. Second, the sensitivity analysis regarding patch size and radius explains the model's reliance on local granularity. The results indicate a clear "structural sweet spot" (approximately 1000 points or 15-20mm radius). Performance degrades with smaller patches (insufficient context to orient the landmark) and with larger patches (introduction of irrelevant noise or background geometry). This demonstrates that PAL-Net's decision-making is driven by a balanced integration of immediate local surface topology and the broader global arrangement of facial features. It is worth noting that while the quantitative differences between these configurations are relatively small, the consistent trend across cross-validation folds reinforces the need for both global attention and optimized local context to minimize outliers and achieve robust anatomical alignment.

\section{Limitations, Conclusions and Future Work}

The proposed pipeline, integrating local patch-based points with global attention, demonstrated high accuracy, stability, and computational efficiency. PAL-Net consistently outperformed existing methods on both point-wise and distance-wise evaluations, achieving an average localization error of 3.686 mm on the LAFAS dataset, with a distance preservation error of 2.822 mm. This performance profile, achieved despite data scarcity supports the model's feasibility as a high-throughput, reliable tool for 3D facial analysis. The lightweight nature of PAL-Net, requiring only 2.5 GiB of GPU memory and less than 4 seconds per training epoch, positions it as a viable solution for real-time clinical and research applications, including facial asymmetry analysis, pre- and post-operative planning, and growth assessment.

Although the proposed PAL-Net framework achieves high accuracy and computational efficiency in anatomical landmark localization, several limitations must be considered. The model was trained on a relatively small dataset (214 subjects) from a single laboratory (LAFAS), which may limit its generalizability to broader populations or pathological cases, despite 5-fold cross-validation. Performance also declined in regions with incomplete or noisy geometry, such as the ears, hairline, and jawline, due to occlusions and surface artifacts inherent to stereo-photogrammetry, affecting both manual and automated annotations. Furthermore, the current pipeline is strictly dependent on the quality of the initial rigid registration of facial models to a shared reference frame. This requirement introduces a structural limitation, as misalignments or registration inaccuracies can negatively affect subsequent landmark approximation and prediction. As a result, the model lacks inherent invariance to rotation and translation. Another limitation lies in the initial landmark approximation based on averaged annotations from the training set. While effective for general structures, this introduces bias in variable regions and may not generalize to atypical facial geometries. To address this, future work will extend the framework to pathological cases where landmarks may deviate significantly from normative patterns. Building a model capable of adapting to such variability is essential for expanding clinical applicability, particularly in cases involving craniofacial syndromes or surgical outcomes. 

Future work will further focus on addressing the mentioned limitations by exploring model architectures that are intrinsically equivariant or invariant to rigid transformations. The integration of geometric learning techniques that handle partial, low-quality, or noisy data may further improve performance in challenging facial regions, such as corrupted points around the ears. Another promising direction for future research is to leverage the predicted soft-tissue landmarks to estimate corresponding skeletal landmarks visible in Computed Tomography (CT) scans \citep{serafin_accuracy_2023}. Establishing reliable mappings between surface-based anatomical landmarks and their underlying skeletal counterparts could enable indirect estimation of craniofacial skeletal points without additional imaging. This approach has the potential to reduce patient exposure to ionizing radiation by minimizing reliance on CT solely for landmark localization, thereby contributing to safer, more efficient clinical workflows.

\section*{Acknowledgments}

Special thanks to Marco Farronato for his helpful feedback and proofreading support during the writing of this manuscript.

Funding: The present research was partially funded by the University of Milan under the “My First SEED Grant” fund, DM 737/2021 MUR (Project: DIAERESES - PSR\_LINEA3\/ Piano di sviluppo di ricerca - Bando SoE-SEED- Linea 3).

\clearpage
\appendix
\section{Per Landmark Model Comparison}

\begin{figure}[!h]
\centering
\includegraphics[width=1\linewidth]{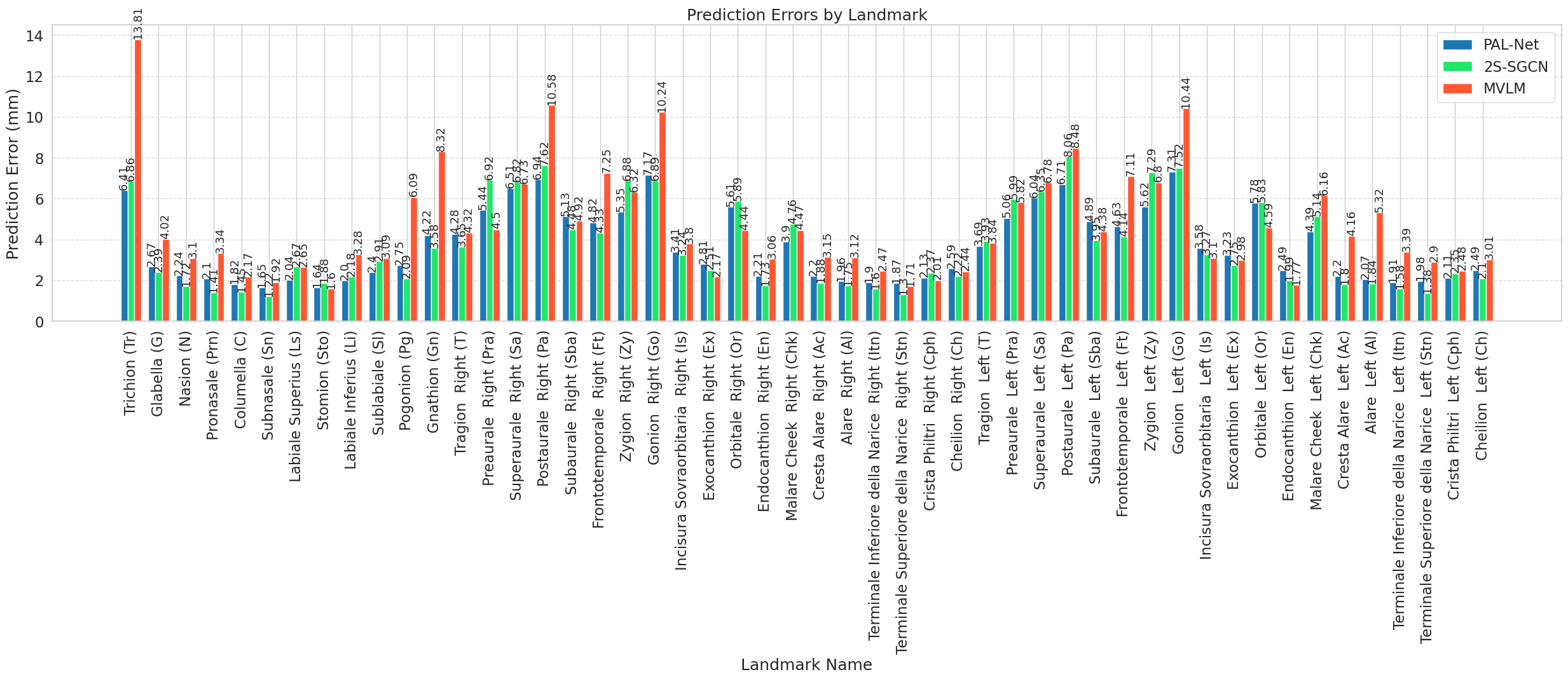}
\caption{ Localization error for each of the 50 landmarks in the LAFAS dataset, comparison of PAL-Net with existing methods. \label{fig:models_comparison_per_landmark} }
\end{figure}

\begin{figure}[!h]
\centering
\includegraphics[width=0.9\linewidth]{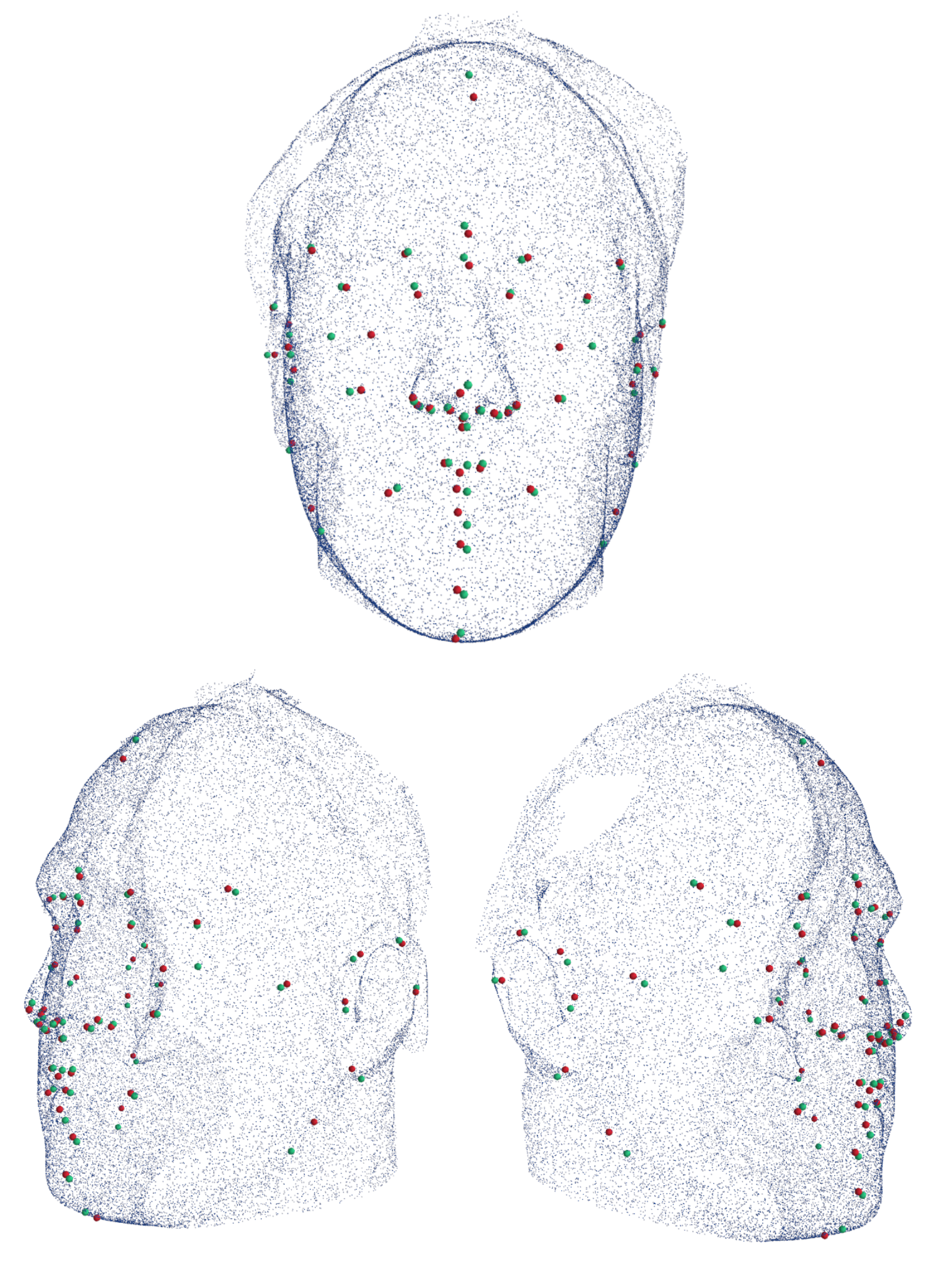}
\caption{\label{fig:model_prediction}Visualization of anatomical landmarks predicted by PAL-Net (red) compared to ground truth annotations (green) on a 3D facial mesh, shown from frontal and lateral views. }
\end{figure}

\clearpage
During the preparation of this work the author(s) used GPT 4o model in order to clarify and enhance the flow of the article for readers. After using this tool/service, the author(s) reviewed and edited the content as needed and take(s) full responsibility for the content of the publication.

\bibliographystyle{elsarticle-num-names}
\bibliography{refrences}
\end{document}